\documentclass{article}
\usepackage{spconf,amsmath,graphicx}

\usepackage[utf8]{inputenc} 
\usepackage[T1]{fontenc}    
\usepackage{hyperref}       
\usepackage{url}            
\usepackage{booktabs}       
\usepackage{amsfonts}       
\usepackage{nicefrac}       
\usepackage{microtype}      
\usepackage{xcolor}         
\usepackage{threeparttable}
\usepackage{makecell}
\usepackage{adjustbox}
\usepackage{multirow}
\usepackage{mathtools}
\usepackage{amsmath}
\usepackage{caption}
\usepackage{subcaption}
\usepackage{lipsum}
\usepackage{wrapfig}
\usepackage{graphicx}
\usepackage{footnote}
\usepackage{easyReview}
\makesavenoteenv{tabular}

\newcommand{\R}{\mathbb{ R}}
\DeclareMathOperator*{\argmin}{argmin}
\title{Text-To-ECG: 12-lead electrocardiogram synthesis \\conditioned on clinical text reports}
\name{Hyunseung Chung$^1$, Jiho Kim$^1$, Joon-myoung Kwon$^{2,4}$, Ki-Hyun Jeon$^3$, Min Sung Lee$^2$, Edward Choi$^1$
\thanks{This work was supported by Medical AI Inc. and Institute of Information \& communications Technology Planning \& Evaluation (IITP) grant (No.2019-0-00075),  Korea Medical Device Development Fund grant (Project Number: 1711138160, KMDF\_PR\_20200901\_0097), and National Research Foundation of Korea (NRF) grant (NRF-2020H1D3A2A03100945) funded by the Korea government(MSIT, MOTIE, MOHW, MFDS).}
}
\address{
  $^1$Kim Jaechul Graduate School of AI, KAIST, Daejeon, Korea\\
  $^2$Medical AI Inc., Seoul, Korea\\
  $^3$Department of Internal Medicine, Seoul National University Bundang Hospital, Seongnam, Korea\\ 
  $^4$Critical Care and Emergency Medicine Department, Sejong Hospital, Incheon, Korea
}
\begin{document}
\ninept
\maketitle
\begin{abstract}
Electrocardiogram (ECG) synthesis is the area of research focused on generating realistic synthetic ECG signals for medical use without concerns over annotation costs or clinical data privacy restrictions.
Traditional ECG generation models consider a single ECG lead and utilize GAN-based generative models. 
These models can only generate single lead samples and require separate training for each diagnosis class.
The diagnosis classes of ECGs are insufficient to capture the intricate differences between ECGs depending on various features (e.g. patient demographic details, co-existing diagnosis classes, etc.).
To alleviate these challenges, we present a text-to-ECG task, in which textual inputs are used to produce ECG outputs.
Then we propose Auto-TTE, an autoregressive generative model conditioned on clinical text reports to synthesize 12-lead ECGs, for the first time to our knowledge.
We compare the performance of our model with other representative models in text-to-speech and text-to-image.
Experimental results show the superiority of our model in various quantitative evaluations and qualitative analysis. Finally, we conduct a user study with three board-certified cardiologists to confirm the fidelity and semantic alignment of generated samples. our code will be available at \href{https://github.com/TClife/text_to_ecg}{https://github.com/TClife/text{\_}to{\_}ecg} 
\end{abstract}


\section{Introduction}
According to the global health mortality analysis \cite{roth2018global}, heart diseases are ranked among the highest causes of death.
In order to prevent or detect symptoms of heart disease in the early stages, clinicians measure electrocardiograms (ECG), which is a non-invasive diagnostic tool to check a patient's heart rhythm and electrical activity.
Interpreting ECGs of patients periodically, however, is a huge burden for a limited number of cardiologists working day in and day out.
Therefore, many attempts have been made in developing ECG arrhythmia classifiers \cite{mousavi2019inter}.
Unfortunately, data privacy restrictions \cite{adib2021synthetic} and annotation costs \cite{chen2022me} limit ECG data usage for training automatic ECG classifiers.
Thus, an important area of research in the medical domain is to generate realistic ECG signals, addressing these issues. 

There is rapid progress in generative models in various domains such as images \cite{esser2021taming, ramesh2021zero}, text \cite{brown2020language}, and speech \cite{shen2018natural, popov2021grad}. The main objective of each area of research is to improve fidelity and diversity of generated samples. This is no exception in the domain of ECG synthesis. In \cite{adib2021synthetic}, various Generative Adversarial Network (GAN) architectures were used to generate ECGs. Then, \cite{golany2020simgans} introduced a system of Ordinary Differential Equations (ODE) representing heart dynamics to create ECG samples. More recently, \cite{chen2022me} introduced ECG generation model conditioned on heart diseases to improve synthesis results.
However, there are still limitations to previous methods because most do not consider multi-view ECGs or other details in text reports.

We propose text-to-ECG generation, in which realistic 12-lead ECGs are generated conditioned on clinical text reports. This improves on the previous ECG synthesis model conditioned on heart diseases by conditioning on full text reports which contain other relevant details of how the ECG is formed. For example, a full text report of an ECG labeled as ``Left Bundle Branch Block'' may be ``Left Bundle Branch Block, commonly due to ischaemic heart disease.'' Ischaemic heart disease is a term given to heart problems caused when arteries are narrowed, and considering such detail during ECG generation will only improve the quality and realism. 

Although there are no existing previous works in the text-to-ECG domain, the text-to-speech domain can be considered most relevant to our work because the objective is to generate time-series data from text inputs. Tacotron 2 \cite{shen2018natural} is considered a representative autoregressive text-to-speech model in speech synthesis, surpassing all of prior works at the time of release. However, Tacotron 2 outputs mel-spectrograms, and therefore must be paired with a vocoder to generate raw audio samples. The most basic form of vocoder is the fast Griffin-Lim algorithm \cite{perraudin2013fast}, which approximates magnitude spectrogram inversion by alternating forward and inverse Short-time Fourier transform (STFT) operations. Then, the Hifi-GAN \cite{kong2020hifi} achieved state-of-the-art vocoder performance. Recently, diffusion probabilistic model \cite{popov2021grad} is introduced as a novel architecture for TTS.       

\begin{figure*}[!t]
    \centering
    \includegraphics[width=2.0\columnwidth, height=7cm]{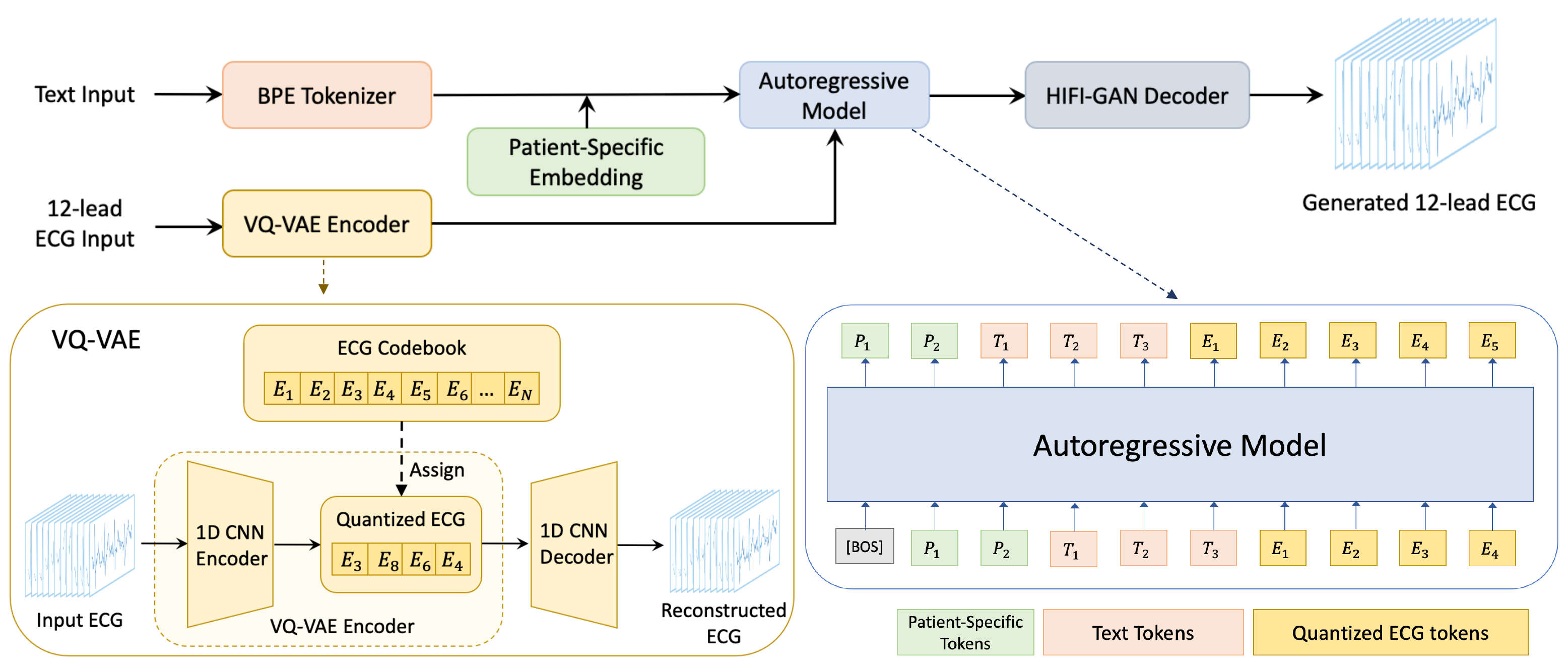}
    \caption{Overall framework of the proposed architecture. There is a BPE Tokenizer (Output can be concatenated with patient-specific embedding), a pre-trained VQ-VAE encoder that quantizes the ECG signal, an autoregressive model that generates quantized ECG tokens from text tokens, and a pre-trained HiFi-GAN decoder that transforms quantized ECG tokens into a raw ECG signal.}
\label{model}
\vspace{-3mm}
\end{figure*}

In this paper, we present Auto-TTE, the first autoregressive text guided 12-lead ECG generation model.
This model is a two-phase network, where in the first phase raw ECG signals are quantized to a sequence of tokens using Vector Quantized-Variational AutoEncoder (VQ-VAE) \cite{van2017neural}.
The model encodes raw ECGs to discrete codes, and reconstructs back to the original raw ECG from discrete representations.
In the second phase, quantized ECG tokens are used as inputs alongside text tokens encoded with BPE \cite{sennrich2015neural} tokenizer.
This single stream of text and ECG tokens are trained autoregressively with a Transformer-decoder model \cite{vaswani2017attention}.  
We present various evaluation methods focusing on different generative aspects of the models such as fidelity and text-ECG alignment. Also, we conduct a user study with three board-certified cardiologists. Our experimental results show our proposed model outperforms other baseline models.  

\section{Model Architecture}
The overall architecture of our proposed framework is shown in Fig.\ref{model}. The architecture consists of three parts: VQ-VAE encoder, Autoregressive Transformer model, and HiFi-GAN decoder.      
  
\subsection{VQ-VAE Encoder}
Although 10 second raw ECG signals contain significantly less timesteps compared to the same 10 second audio signals due to the difference in the sampling rate (500Hz vs. 16kHz), our ECG signals have 10 second ECG segment in 12 different views. Therefore, the amount of information for each ECG signal is multiplied by a factor of 12.
Models like the Variational Autoencoder \cite{kingma2013auto} is commonly used for compression, but the learned continuous latent representation cannot be used as inputs to a Transformer model. To this end, we use the VQ-VAE, which is able to compress the raw ECG signals and quantize them to discrete latent representations.    

VQ-VAE training process contains three parts, which are encoder, a codebook, and decoder. The encoder $E$ consists of four convolutional layers to downsample the 12-lead ECG signal. The codebook $C$, also defined as the latent embedding space, consists of code vectors ${c_k}\in \R^{K\times d}$, where $K$ represents the codebook size and $d$ the dimension of each code vector $c_k, k \in 1,2,...,K$.
Given a raw ECG signal $x \in \R^{L\times T}$ with $L$ leads and $T$ timesteps, $x$ goes through the encoder to produce output $\hat{l} = E_c(x) \in \R^{T^{\prime}\times d}$, where $T^{\prime}$ is the reduced time dimension after downsampling. The output after quantization $E_q(.)$ process is as follows:  
\begin{equation}
    E_q(\hat{l}) := (\argmin_{c_k \in C}\lVert \hat{l}_i - c_{k} \rVert_2^2 \: \text{for all $i$ in $T^{\prime}$})\nonumber
\end{equation}
Then, the decoder $U$ reconstructs the input $\hat{x} = U(E_q(\hat{l}))$. The entire process is trained with the following loss function: 
\begin{equation}
    L_{VQ} = \lVert x - \hat{x}\rVert_2^2 + \lVert \text{sg}[E_c(x)] - E_q(\hat{l})\rVert_2^2 + \lVert \text{sg}[E_q(\hat{l})] - E_c(x)\rVert_2^2 \nonumber
\end{equation}
where sg stands for stop-gradient, which is an identity during the forward pass and zero gradient during the backward propagation.

\subsection{Autoregressive Transformer-decoder model}
Given a text-ECG pair, the text inputs pass through the BPE Tokenizer to produce a sequence of subword indices, or text tokens.
Also, patient-specific information is used to produce each of the patient-specific tokens.
We consider only age and gender of patients, but any patient-specific demographic information can be used for this process.
Patient-specific tokens are prepended to text tokens as a single sequence, and a lookup table produces a token embedding, $u$.
Similarly, after fully training the VQ-VAE, the encoder quantizes a 12-lead ECG signal as a sequence of code indices $E_q(\hat{l})$ with the codebook embeddings. These indices are considered as quantized ECG tokens, and also goes through a lookup function to produce ECG token embeddings, $v$. Then, both $u$ and $v$ embeddings are concatenated to create $S=u_1, u_2, ..., u_{T_a}, v_1, v_2, ..., v_{T_b}$ where $T_a$ represents the length of patient-specific and text token embeddings, and $T_b$ represents the length of ECG token embeddings. Finally, the embeddings are used to train a Transformer-decoder model with the cross-entropy loss function parameterized by $\theta$ as follows: 
\begin{equation}
    L_{T} = -\sum_{i=1}^{M} \sum_{j=1}^{T_b}\log{p_{\theta}}(v_{j}^{i}|u_{1}^{1},...,u_{T_a}^{i},v_{1}^{1},...,v_{j-1}^{i})\nonumber   
\end{equation}
where $M$ represents the number of samples. Each ECG token can attend to any of the previous text tokens, and standard causal mask is used for text-to-text and ECG-to-ECG attention masks. The final tokens are selected with argmax sampling. 

\begin{table*} [!t]
    \caption{
    The AUROC column represents Area Under the Receiver Operating Characteristic Curve (AUROC) scores of samples generated with each model by using text prompts and six different diagnosis classes as text inputs. For all tables, $\ast$ refers to p-value < 0.05 (statistically significant) compared to ``Auto-TTE''.  
    }
    \label{evaluation}
    \vspace{-3mm}
    \begin{threeparttable}
    \centering
    \begin{tabular}{l|c c c c c c |c|c} \Xhline{3\arrayrulewidth}
        \multicolumn{1}{c|}{\multirow{2}{*}{\textbf{Model}}}
        &\multicolumn{7}{c|}{\textbf{Classification Results (AUROC)}}
        &\multicolumn{1}{c}{\multirow{2}{*}{\textbf{CLIP score}}}
        \\
        $\empty$
        &\textbf{AFIB} &\textbf{Bradycardia} &\textbf{LBBB} &\textbf{Normal ECG} &\textbf{RBBB} &\textbf{Tachycardia } &\textbf{Average}
        &\textbf{$\empty$} 
        \\ \hline
            Ground Truth
            &$0.995$ &$0.983$&$0.987$&$0.990$&$0.977$&$0.981$& $0.986$ & $0.7010$   \\ \hline 
            Diffusion Model &$0.730^{\ast}$ &$0.735^{\ast}$&$0.667^{\dagger\ast}$&$0.669^{\dagger\ast}$&$0.723^{\ast}$&$0.803^{\dagger\ast}$&$0.721$&$0.4711^{\ast}$ \\ 
            Tacotron 2 + Griffin-Lim &$0.664^{\dagger\ast}$ &$0.574^{\dagger\ast}$&$0.633^{\ast}$&$0.610^{\ast}$&$0.603^{\ast}$&$0.644^{\ast}$&$0.621$&$0.3746^{\ast}$  \\
            Tacotron 2 + Hifi-GAN &$0.820$ &$0.822^{\ast}$&$0.846^{\ast}$&$0.811^{\dagger}$&$\textbf{0.894}^{\dagger\ast}$&$0.828^{\dagger\ast}$&$0.834$&$0.6209^{\ast}$  \\\hline
            Auto-TTE &$\textbf{0.824}$ &$\textbf{0.848}$&$\textbf{0.905}$&$\textbf{0.839}^{\dagger}$&$0.841$&$\textbf{0.886}$&$\textbf{0.857}$&$\textbf{0.6444}$   \\ 
       \Xhline{3\arrayrulewidth}
    \end{tabular}
    \begin{tablenotes}[flushleft]
	 \item  {\footnotesize$\dagger$: standard deviation > $0.02$, $\ast$: p value < $0.05$}
	\end{tablenotes}
    \end{threeparttable}
    \vspace{-3mm}
\end{table*}


\subsection{Hifi-GAN Decoder}
A fully trained VQ-VAE decoder $U$ can be used to decode the Transformer output to a 12-lead ECG signal.
However, a decoder that properly models different frequency components and periodic patterns of ECG signals can generate higher quality samples. Therefore, we utilize the Hifi-GAN \textbf{}\cite{kong2020hifi} architecture, which contains the aforementioned attributes in two discriminators with different training objectives. 
The first discriminator is the Multi-Scale Discriminator \cite{kumar2019melgan} that effectively learns different frequency components of signals through variation in scales.
The second discriminator is the Multi-Period Discriminator, which operates convolution process in periodic variations. Distinct periodic features are extracted by considering different periodic segments of ECG signals.

\begin{figure}[!t]
    \includegraphics[width=1\columnwidth]{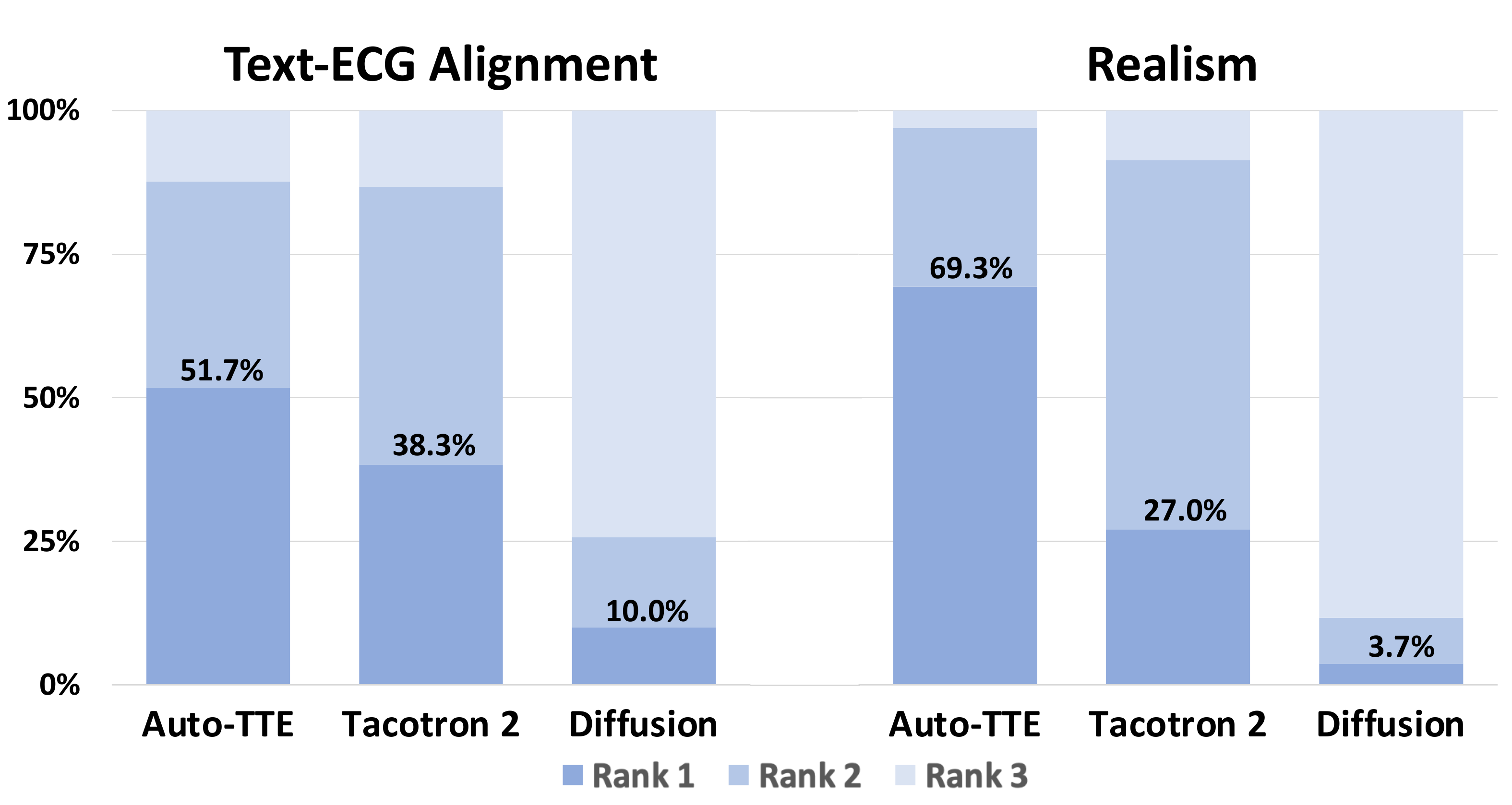}
    \caption{Human evaluation of Auto-TTE (ours) vs Tacoton2 + Hifi-GAN and Diffusion model. Percentages of rank 1 are displayed.}
    \vspace{-2mm}
    \label{user_study}
    \vspace{-2mm}
\end{figure}
The difference between the original Hifi-GAN model and our Hifi-GAN decoder is that the original model used Hifi-GAN to produce raw audio waveforms from mel-spectrogram inputs, while we produce raw ECG signals from quantized ECG tokens.
Our generator is a fully convolutional neural network consisting of several transposed convolutions to upsample the quantized ECG token inputs to the raw 12-lead ECG signal outputs.
We will denote our generator as $G$ and both our discriminators as a single discriminator $D$. Then, the training objective follows LS-GAN \cite{mao2017least}, which uses least squares instead of binary cross-entropy loss.
The loss is defined as:
\begin{equation}
    \begin{split}
    \mathcal{L}&_{adv}(D)=
         \mathbb{E}_{(x,E_q(\hat{l}))}\bigg[(D(x)-1)^2+
         (D(G(E_q(\hat{l}))))^2\bigg]\nonumber\\
    \end{split}
\end{equation}
\begin{equation}
    \begin{split}
    \mathcal{L}_{adv}(G)&=\mathbb{E}_s\bigg[( D(G(E_q(\hat{l})))-1)^2\bigg]\nonumber\\
    \end{split}
\end{equation}
The discriminator is trained to classify samples synthesized from the generator to zero, and ground-truth samples to one. Additionally, the spectrogram loss and feature matching loss are presented as:

\begin{equation}
    \begin{split}
    \mathcal{L}_{spec}(G) = \mathbb{E}\bigg[||{spec}(x)-{spec}(G(E_q(\hat{l})))||_{1}\bigg]\nonumber
    \end{split}
\end{equation}
\begin{equation}
    {L}_{fm}(G) =\mathbb{E}\bigg[\sum_{i=0}^{Z-1}\frac{1}{N_i}\lVert\nonumber
    D^{i}(x)-D^{i}(G(E_q(\hat{l})))\rVert_1\bigg]
\end{equation}
where the $spec(.)$ function transforms the input signal into a spectrogram using STFT, $Z$ represents the number of layers in the discriminator, and $D^{i}$ and $N_{i}$ means the features and number of features in i-th discriminator layer. Finally, the final objective is a weighted sum of all three losses.

\section{Experiments}
\subsection{Experimental Settings}
We conduct experiments on the PTB-XL dataset \cite{wagner2020ptb} and Sejong dataset. The publicly available PTB-XL dataset consists of 21,833 text-ECG pair samples. Sejong dataset is a private dataset of 37,329 text-ECG pairs provided by Sejong General Hospital.
All ECG samples in both datasets are 12-lead with 10-second segments sampled in 500Hz.
After combining the two datasets, we randomly split the dataset into train (80\%), validation (10\%) and test (10\%) sets. All experiments were conducted with 4 \textbf{RTX A6000} GPUs. 

\begin{table}[!t]
    \caption{FID score, Precision, and Recall. $\ast$ refers to p-value < 0.05 (statistically significant) compared to Auto-TTE.}
    \label{evaluation2}
    \vspace{-3mm}
    \begin{threeparttable}
    \centering
     \resizebox{1\columnwidth}{!}{
    \begin{tabular}{l|c|c|c} \Xhline{3\arrayrulewidth}
       
     \textbf{Model} &\textbf{FID($\downarrow$)} &\textbf{Precision}  &\textbf{Recall}    \\  \hline
            Diffusion Model &$16.46^{\ast}$  &$0.896^{\ast}$  &$0.598^{\ast}$\\ 
            Tacotron 2 + Griffin-Lim &$22.93^{\ast}$  &$0.887^{\ast}$  &$0.423^{\ast}$\\
            Tacotron 2 + Hifi-GAN &$9.18^{\ast}$  &$0.934$  &$0.646^{\ast}$\\ \hline
            Auto-TTE &$\textbf{7.17}^{\dagger}$  &$\textbf{0.946}$  &$\textbf{0.842}$\\
            
        \Xhline{3\arrayrulewidth}
    \end{tabular}

    }
    \begin{tablenotes}[flushleft]
	 \item  {\footnotesize$\dagger$: standard deviation > $0.2$, $\ast$: p value < $0.05$}
	\end{tablenotes}
    \end{threeparttable}
    \vspace{-3mm}
\end{table}

\subsection{Training Setup}
We train all models with a batch size of 64. For Tacotron 2 and Hifi-GAN, we use linear spectrogram which is transformed from raw ECG signals with STFT using 32 window size, 8 hop size, and 32 FFT.
We also use the V1 model for Hifi-GAN, which is the large parameter model with initial channel of 512.
For Hifi-GAN training, we utilized 1, 2, and 45 for least squares, spectrogram, and feature matching loss weights, respectively.
Also, we used AdamW optimizer \cite{loshchilov2017decoupled} with $\beta_{1}=0.8, \beta_{2}=0.99$ and weight decay of $\lambda=0.01$. The Transformer-decoder model contained 12 attention layers, 8 attention heads, and per-head dimension 64.  
For Tacotron 2, VQ-VAE, and autoregressive Transformer-decoder model training, we optimized the models using Adam \cite{kingma2014adam} optimizer, and learning rate of $2 \times 10^{-4}$.
The BPE tokenizer encoded text using at most 128 tokens, and a vocabulary size of 3,000. The VQ-VAE compressed $12 \times 5000$ ECG signals to 312 tokens, reduced by a factor of 192 with vocabulary size of 1024.
As an additional baseline, we use a diffusion-based model.
After trying numerous variations (\textit{e.g.} generate mel-spectrogram, generate latent representation \cite{rombach2022high}), we used diffusion directly on raw 12-lead ECG signals, which showed the best performance.
Our U-net architecture is similar to the architecture in \cite{ho2020denoising}, except that we used 12 channels and 1-D convolutions for ECG data.              

\begin{table}[!t]
    \caption{AUROC and Clip score results for Auto-TTE and ablations.}
    \vspace{-3mm}
    \label{evaluation3}
    \begin{threeparttable}
    \centering
     \resizebox{1\columnwidth}{!}{
    \begin{tabular}{l|c|c} \Xhline{3\arrayrulewidth}
       
     \textbf{Model} &\textbf{Average AUROC} &\textbf{CLIP score}   \\  \hline
            No PSE \& Hifi-GAN decoder &$0.829$  &$0.5833^{\ast}$\\
            No PSE &$0.832^{\ast}$  &$0.5908^{\ast}$\\
            No Hifi-GAN decoder &$0.817^{\dagger}$  &$0.6104^{\ast}$\\ \hline
            Auto-TTE &$\textbf{0.857}$  &$\textbf{0.6444}$\\
            
        \Xhline{3\arrayrulewidth}
    \end{tabular}
    }
    \begin{tablenotes}[flushleft]
	 \item  {\footnotesize$\dagger$: standard deviation > $0.02$, $\ast$: p value < $0.05$}
	\end{tablenotes}
    \end{threeparttable}
\end{table}

\begin{table}[!t]
    \caption{FID score, Precision, and Recall for Auto-TTE and ablations.}
    \vspace{-3mm}
    \label{evaluation4}
    \begin{threeparttable}
    \centering
     \resizebox{1\columnwidth}{!}{
    \begin{tabular}{l|c|c|c} \Xhline{3\arrayrulewidth}
       
     \textbf{Model} &\textbf{FID($\downarrow$)} &\textbf{Precision}  &\textbf{Recall}    \\  \hline
            No PSE \& Hifi-GAN decoder &$9.27^{\dagger\ast}$  &$0.912$  &$0.812$\\
            No PSE &$8.76^{\ast}$  &$0.911^{\ast}$  &$0.832$\\
            No Hifi-GAN decoder &$7.21^{\dagger}$  &$0.926$  &$0.812^{\ast}$\\ \hline
            Auto-TTE &$\textbf{7.17}^{\dagger}$  &$\textbf{0.946}$  &$\textbf{0.842}$\\
            
        \Xhline{3\arrayrulewidth}
    \end{tabular}
    }
    \begin{tablenotes}[flushleft]
	 \item  {\footnotesize$\dagger$: standard deviation > $0.2$, $\ast$: p value < $0.05$}
	\end{tablenotes}
    \end{threeparttable}
    \vspace{-3mm}
\end{table}


\subsection{Quantitative Results}
\label{quantitative}
We conduct various evaluations to compare our model against other baseline models: Diffusion model, Tacotron 2 + Griffin-Lim, and Tacotron 2 + Hifi-GAN.

The first evaluation is conducted to test how well the generated ECGs can capture representative cardiac diagnoses when they are given as textual input. We first collect most frequent textual prompts with each diagnosis in our dataset, which is determined with a Word N-gram of $N$ between 4 and 10. Then, we use the corresponding ECGs to train a multi-label classifier with ResNet-34 \cite{he2016deep} architecture and 1-D convolutions. Finally, we generate 128 ECG samples for each diagnosis with the collected textual prompts. The classification results of different models are shown in Table \ref{evaluation}.

The second evaluation is conducted to assess text-ECG alignment performance, and is based on Contrastive Language-Image Pre-training (CLIP) \cite{radford2021learning} method. 
Instead of training an image encoder, we train an ECG encoder jointly with a text encoder to predict correct pairings of ECG and text.
We then generate ECGs based on full text reports in the test set with each model.
During test time, the CLIP score between a text report and a generated ECG is evaluated.
The positive score results of this evaluation is shown in Table \ref{evaluation}.

The third method is conducted to evaluate fidelity of generated ECGs. It is the Fréchet inception distance (FID) score for ECGs based on a self-supervised learning model introduced in \cite{oh2022lead}, which is pre-trained in Physionet 2021 dataset \cite{reyna2021will}.
We modify the last layer of the pre-trained model to output a 64-dimensional feature vector.
The generated ECGs from the CLIP score evaluation and ground-truth test set ECGs are both passed through the pre-trained model to produce feature vectors for FID score evaluation. Afterwards, precision and recall values are also calculated following \cite{kynkaanniemi2019improved}.
The results are shown in Table \ref{evaluation2}. For all evaluations, we conduct 10 random multiple-bootstrap experiments and report mean and standard deviations.
Also, we perform statistical hypothesis test by conducting independent t-test with a significance value of 0.05 to identify statistically significant pairwise difference between Auto-TTE and other models.
Auto-TTE has the best results in most evaluation methods.    

We also conduct a human evaluation with three board-certified cardiologists to compare our approach with Tacotron 2 and Diffusion. 
Given a full ECG text report, the cardiologists compared the generated ECGs from each model and ranked them for text-ECG alignment (how accurately the ECGs align with the text report) and realism (how realistic the ECGs themselves are). We present average rankings of 100 samples from each model, as shown in Figure \ref{user_study}. The ECGs generated with Auto-TTE were ranked the highest for both text-ECG alignment and realism, followed by Tacotron 2 and Diffusion.        

\subsection{Qualitative Analysis}
\begin{figure}[!t]
\centering
\subfloat[ECGs generated with different models and same text]{%
  \includegraphics[clip,width=\columnwidth]{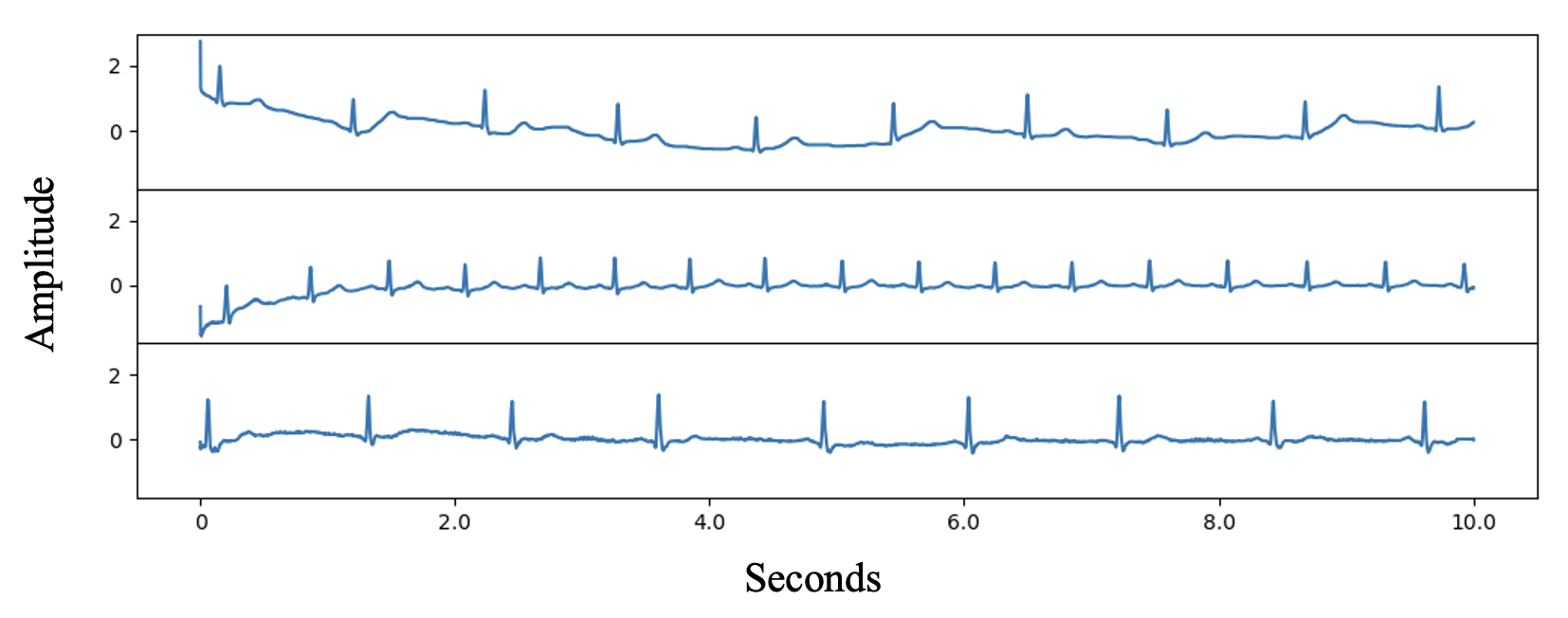}%
}

\subfloat[ECGs generated with Auto-TTE and different texts]{%
  \includegraphics[clip,width=\columnwidth]{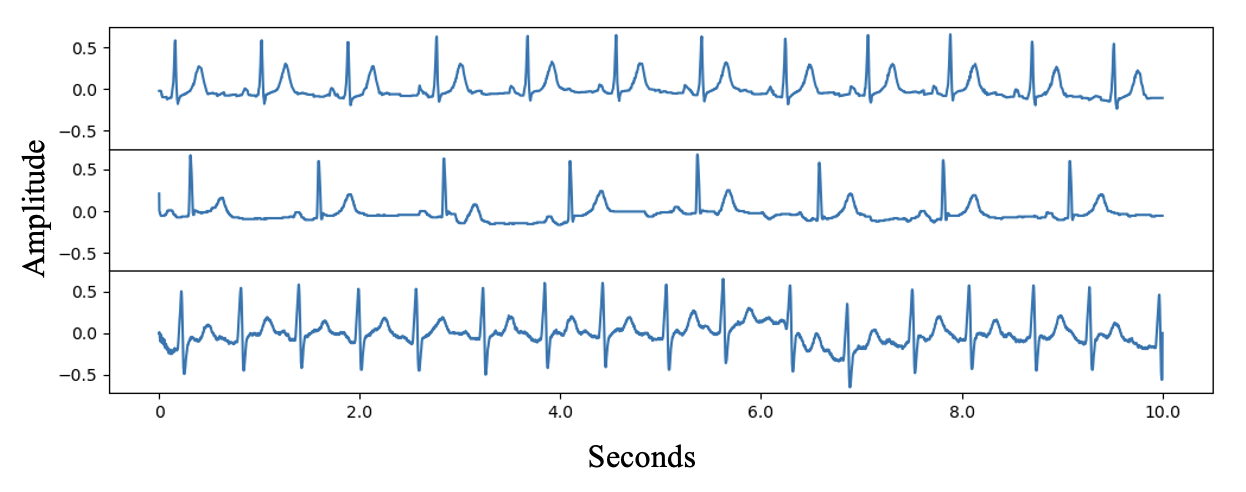}%
}
\vspace{-3mm}
\caption{Figure (a) shows the ``V6'' lead of generated ECG samples with Auto-TTE, Tacotron 2 + Hifi-GAN, and Diffusion models with text ``Baseline Wander in Lead(s) V6''. Figure (b) shows lead II and are generated with Auto-TTE. Texts used are ``V-rate 50-99'', ``V-rate<50'', and ``V-rate>99''. All orders are in top-to-bottom.}
\vspace{-3mm}
\label{qualitative}
\end{figure}

To examine the ability of our model to generate accurate ECGs conditioned on diverse details in text reports, we first use the most commonly occurring artifact in the ECG: baseline wander, which is a baseline drift effect. We compare the generation result from using a text input that indicates baseline wander in a specific lead.
Also, we compare generated ECGs by Auto-TTE using text inputs that alter ventricular rate (v-rate), which is heart rate per minute. Results are shown in Figure \ref{qualitative}. 

\subsection{Ablation Study}
We conduct ablation study for Auto-TTE, and evaluate in the same methods described in Section \ref{quantitative}. In Table \ref{evaluation3}, the average AUROC and CLIP score is the highest for Auto-TTE, and largely decreases with the absence of Hifi-GAN decoder and PSE, which refers to Patient-Specific Embedding. Here, ``No Hifi-GAN decoder'' substitutes Hifi-GAN decoder with pre-trained VQ-VAE decoder. In Table \ref{evaluation4}, similar trends are also shown for FID, precision, and recall values.

\section{Conclusion}
In this work, we propose a novel text-to-ECG task and a text-to-ECG generation model. Our model, Auto-TTE, first quantizes ECGs into discrete representations, and performs autoregressive Transformer model training with text and patient-specific embeddings. Then, the decoder synthesizes 12-lead ECG signals with high realism and semantic alignment to full text reports. By rigorously evaluating our model on various evaluation methods and conducting a user study with board-certified cardiologists, we demonstrate the superior performance of Auto-TTE compared to other baseline models.          

\newpage

\bibliographystyle{IEEEbib}
\bibliography{strings,refs}

\begin{thebibliography}{10}

\bibitem{roth2018global}
Gregory~A Roth, Degu Abate, Kalkidan~Hassen Abate, Solomon~M Abay, Cristiana
  Abbafati, Nooshin Abbasi, Hedayat Abbastabar, Foad Abd-Allah, Jemal Abdela,
  Ahmed Abdelalim, et~al.,
\newblock ``Global, regional, and national age-sex-specific mortality for 282
  causes of death in 195 countries and territories, 1980--2017,''
\newblock {\em The Lancet}, vol. 392, pp. 1736--1788, 2018.

\bibitem{mousavi2019inter}
Sajad Mousavi and Fatemeh Afghah,
\newblock ``Inter-and intra-patient ecg heartbeat classification for arrhythmia
  detection: a sequence to sequence deep learning approach,''
\newblock in {\em ICASSP 2019-2019 IEEE International Conference on Acoustics,
  Speech and Signal Processing (ICASSP)}. IEEE, 2019, pp. 1308--1312.

\bibitem{adib2021synthetic}
Edmond Adib, Fatemeh Afghah, and John~J Prevost,
\newblock ``Synthetic ecg signal generation using generative neural networks,''
\newblock {\em arXiv preprint arXiv:2112.03268}, 2021.

\bibitem{chen2022me}
Jintai Chen, Kuanlun Liao, Kun Wei, Haochao Ying, Danny~Z Chen, and Jian Wu,
\newblock ``Me-gan: Learning panoptic electrocardio representations for
  multi-view ecg synthesis conditioned on heart diseases,''
\newblock in {\em International Conference on Machine Learning}. PMLR, 2022,
  pp. 3360--3370.

\bibitem{esser2021taming}
Patrick Esser, Robin Rombach, and Bjorn Ommer,
\newblock ``Taming transformers for high-resolution image synthesis,''
\newblock in {\em Proceedings of the IEEE/CVF conference on computer vision and
  pattern recognition}, 2021, pp. 12873--12883.

\bibitem{ramesh2021zero}
Aditya Ramesh, Mikhail Pavlov, Gabriel Goh, Scott Gray, Chelsea Voss, Alec
  Radford, Mark Chen, and Ilya Sutskever,
\newblock ``Zero-shot text-to-image generation,''
\newblock in {\em International Conference on Machine Learning}. PMLR, 2021,
  pp. 8821--8831.

\bibitem{brown2020language}
Tom Brown, Benjamin Mann, Nick Ryder, Melanie Subbiah, Jared~D Kaplan, Prafulla
  Dhariwal, Arvind Neelakantan, Pranav Shyam, Girish Sastry, Amanda Askell,
  et~al.,
\newblock ``Language models are few-shot learners,''
\newblock {\em Advances in neural information processing systems}, vol. 33, pp.
  1877--1901, 2020.

\bibitem{shen2018natural}
Jonathan Shen, Ruoming Pang, Ron~J Weiss, Mike Schuster, Navdeep Jaitly,
  Zongheng Yang, Zhifeng Chen, Yu~Zhang, Yuxuan Wang, Rj~Skerrv-Ryan, et~al.,
\newblock ``Natural tts synthesis by conditioning wavenet on mel spectrogram
  predictions,''
\newblock in {\em 2018 IEEE international conference on acoustics, speech and
  signal processing (ICASSP)}. IEEE, 2018, pp. 4779--4783.

\bibitem{popov2021grad}
Vadim Popov, Ivan Vovk, Vladimir Gogoryan, Tasnima Sadekova, and Mikhail
  Kudinov,
\newblock ``Grad-tts: A diffusion probabilistic model for text-to-speech,''
\newblock in {\em International Conference on Machine Learning}. PMLR, 2021,
  pp. 8599--8608.

\bibitem{golany2020simgans}
Tomer Golany, Kira Radinsky, and Daniel Freedman,
\newblock ``Simgans: Simulator-based generative adversarial networks for ecg
  synthesis to improve deep ecg classification,''
\newblock in {\em International Conference on Machine Learning}. PMLR, 2020,
  pp. 3597--3606.

\bibitem{perraudin2013fast}
Nathana{\"e}l Perraudin, Peter Balazs, and Peter~L S{\o}ndergaard,
\newblock ``A fast griffin-lim algorithm,''
\newblock in {\em 2013 IEEE Workshop on Applications of Signal Processing to
  Audio and Acoustics}. IEEE, 2013, pp. 1--4.

\bibitem{kong2020hifi}
Jungil Kong, Jaehyeon Kim, and Jaekyoung Bae,
\newblock ``Hifi-gan: Generative adversarial networks for efficient and high
  fidelity speech synthesis,''
\newblock {\em Advances in Neural Information Processing Systems}, vol. 33, pp.
  17022--17033, 2020.

\bibitem{van2017neural}
Aaron Van Den~Oord, Oriol Vinyals, et~al.,
\newblock ``Neural discrete representation learning,''
\newblock {\em Advances in neural information processing systems}, vol. 30,
  2017.

\bibitem{sennrich2015neural}
Rico Sennrich, Barry Haddow, and Alexandra Birch,
\newblock ``Neural machine translation of rare words with subword units,''
\newblock {\em arXiv preprint arXiv:1508.07909}, 2015.

\bibitem{vaswani2017attention}
Ashish Vaswani, Noam Shazeer, Niki Parmar, Jakob Uszkoreit, Llion Jones,
  Aidan~N Gomez, {\L}ukasz Kaiser, and Illia Polosukhin,
\newblock ``Attention is all you need,''
\newblock {\em Advances in neural information processing systems}, vol. 30,
  2017.

\bibitem{kingma2013auto}
Diederik~P Kingma and Max Welling,
\newblock ``Auto-encoding variational bayes,''
\newblock {\em arXiv preprint arXiv:1312.6114}, 2013.

\bibitem{kumar2019melgan}
Kundan Kumar, Rithesh Kumar, Thibault de~Boissiere, Lucas Gestin, Wei~Zhen
  Teoh, Jose Sotelo, Alexandre de~Br{\'e}bisson, Yoshua Bengio, and Aaron
  Courville,
\newblock ``Melgan: Generative adversarial networks for conditional waveform
  synthesis,''
\newblock {\em Advances in neural information processing systems}, vol. 32,
  2019.

\bibitem{mao2017least}
Xudong Mao, Qing Li, Haoran Xie, Raymond~YK Lau, Zhen Wang, and Stephen
  Paul~Smolley,
\newblock ``Least squares generative adversarial networks,''
\newblock in {\em Proceedings of the IEEE international conference on computer
  vision}, 2017, pp. 2794--2802.

\bibitem{wagner2020ptb}
Patrick Wagner, Nils Strodthoff, Ralf-Dieter Bousseljot, Dieter Kreiseler,
  Fatima~I Lunze, Wojciech Samek, and Tobias Schaeffter,
\newblock ``Ptb-xl, a large publicly available electrocardiography dataset,''
\newblock {\em Scientific data}, vol. 7, no. 1, pp. 1--15, 2020.

\bibitem{loshchilov2017decoupled}
Ilya Loshchilov and Frank Hutter,
\newblock ``Decoupled weight decay regularization,''
\newblock {\em arXiv preprint arXiv:1711.05101}, 2017.

\bibitem{kingma2014adam}
Diederik~P Kingma and Jimmy Ba,
\newblock ``Adam: A method for stochastic optimization,''
\newblock {\em arXiv preprint arXiv:1412.6980}, 2014.

\bibitem{rombach2022high}
Robin Rombach, Andreas Blattmann, Dominik Lorenz, Patrick Esser, and Bj{\"o}rn
  Ommer,
\newblock ``High-resolution image synthesis with latent diffusion models,''
\newblock in {\em Proceedings of the IEEE/CVF Conference on Computer Vision and
  Pattern Recognition}, 2022, pp. 10684--10695.

\bibitem{ho2020denoising}
Jonathan Ho, Ajay Jain, and Pieter Abbeel,
\newblock ``Denoising diffusion probabilistic models,''
\newblock {\em Advances in Neural Information Processing Systems}, vol. 33, pp.
  6840--6851, 2020.

\bibitem{he2016deep}
Kaiming He, Xiangyu Zhang, Shaoqing Ren, and Jian Sun,
\newblock ``Deep residual learning for image recognition,''
\newblock in {\em Proc. IEEE Conference on Computer Vision and Pattern
  Recognition (CVPR)}, 2016, pp. 770--778.

\bibitem{radford2021learning}
Alec Radford, Jong~Wook Kim, Chris Hallacy, Aditya Ramesh, Gabriel Goh,
  Sandhini Agarwal, Girish Sastry, Amanda Askell, Pamela Mishkin, Jack Clark,
  et~al.,
\newblock ``Learning transferable visual models from natural language
  supervision,''
\newblock in {\em International Conference on Machine Learning}. PMLR, 2021,
  pp. 8748--8763.

\bibitem{oh2022lead}
Jungwoo Oh, Hyunseung Chung, Joon-myoung Kwon, Dong-gyun Hong, and Edward Choi,
\newblock ``Lead-agnostic self-supervised learning for local and global
  representations of electrocardiogram,''
\newblock in {\em Conference on Health, Inference, and Learning}. PMLR, 2022,
  pp. 338--353.

\bibitem{reyna2021will}
Matthew~A Reyna, Nadi Sadr, Erick A~Perez Alday, Annie Gu, Amit~J Shah, Chad
  Robichaux, Ali~Bahrami Rad, Andoni Elola, Salman Seyedi, Sardar Ansari,
  et~al.,
\newblock ``Will two do? varying dimensions in electrocardiography: the
  physionet/computing in cardiology challenge 2021,''
\newblock in {\em 2021 Computing in Cardiology (CinC)}. IEEE, 2021, vol.~48,
  pp. 1--4.

\bibitem{kynkaanniemi2019improved}
Tuomas Kynk{\"a}{\"a}nniemi, Tero Karras, Samuli Laine, Jaakko Lehtinen, and
  Timo Aila,
\newblock ``Improved precision and recall metric for assessing generative
  models,''
\newblock {\em Advances in Neural Information Processing Systems}, vol. 32,
  2019.

\end{thebibliography}

\end{document}